\documentclass[10pt,twocolumn,letterpaper]{article}

\usepackage{cvpr}
\usepackage{times}
\usepackage{epsfig}
\usepackage{graphicx}
\usepackage{amsmath}
\usepackage{amssymb}

\usepackage{graphicx}
\usepackage[margin=1.1in]{geometry}
\usepackage[font=footnotesize, format=plain]{caption}
\usepackage{subcaption}
\usepackage{color}
\usepackage{multirow}
\usepackage{amsfonts}
\usepackage{wrapfig}
\usepackage{dsfont}
\usepackage{enumitem}
\usepackage{array, booktabs}
\usepackage{amsmath}
\usepackage[]{algorithm2e}
\usepackage{url}

\usepackage[pagebackref=true,breaklinks=true,letterpaper=true,colorlinks,bookmarks=false]{hyperref}

\cvprfinalcopy 

\def\cvprPaperID{647} 

\ifcvprfinal\fi
\begin{document}

\newcommand{\mc}[1] {\mathcal{#1}}
\newcommand{\mb}[1] {\mathbf{#1}}
\newcommand{\hdf}[1] {\noindent{\bf #1}}
\newcommand{\hd}[1] {\vspace*{4pt}\noindent{\bf #1}}

\title{Fine-To-Coarse Global Registration of RGB-D Scans }

\author{Maciej Halber\\
Princeton University\\
{\tt\small mhalber@cs.princeton.edu}
\and
Thomas Funkhouser\\
Princeton University\\
{\tt\small funk@cs.princeton.edu}
}

\maketitle


\begin{abstract}

RGB-D scanning of indoor environments is important for many
applications, including real estate, interior design, and virtual
reality.  However, it is still challenging to register RGB-D images
from a hand-held camera over a long video sequence into a globally
consistent 3D model.  Current methods often can lose tracking or drift
and thus fail to reconstruct salient structures in large
environments (e.g., parallel walls in different rooms).  To address
this problem, we propose a ``fine-to-coarse'' global registration
algorithm that leverages robust registrations at finer scales to seed
detection and enforcement of new correspondence and structural
constraints at coarser scales.  To test global registration algorithms, we provide a benchmark with 10,401 manually-clicked point correspondences in 25 scenes from the SUN3D dataset.  During experiments with this benchmark, we find that our fine-to-coarse algorithm registers long RGB-D sequences better than previous methods.

\end{abstract}

\section{Introduction}
\label{sec:introduction}

\begin{figure}[t]
\centering
\includegraphics[width=\columnwidth]{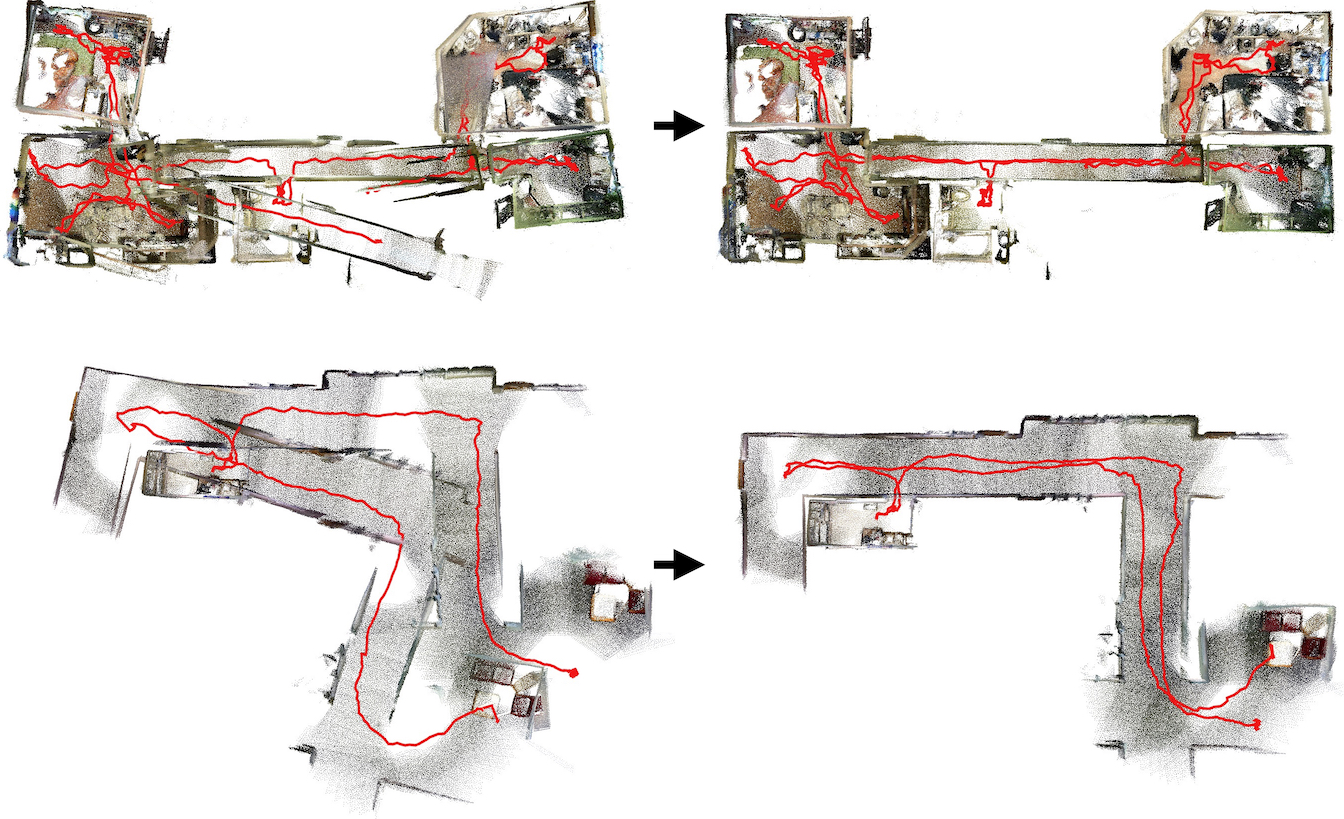}
\captionof{figure}{We present a fine-to-coarse optimization strategy
for globally registering RGB-D scans in indoor environments.  Given
an initial registration (left), our algorithm iteratively detects
and enforces planar structures and feature correspondences at increasing
scales.  In this way, it discovers long-range constraints important
for a globally consistent registration -- e.g., note how opposing walls
are parallel even across different rooms in our results on the right.}
\label{fig:teaser}
\end{figure}

The proliferation of inexpensive RGB-D video cameras allows for easy
scanning of static indoor environments, enabling applications in many domains,
including cultural heritage, real estate and virtual reality.
Motivated by these applications, our goal is to create a method that
takes a sequence of RGB-D images captured with a hand-held camera as
input and produces a globally consistent 3D model as output.  We
would like the algorithm to register images robustly in a wide range
of indoor environments (offices, homes, museums, etc.), execute
off-line within practical computational limits, and work with data
acquired by inexpensive commodity cameras, so that it can be used by
non-experts.

Despite much prior work, it is still difficult to register RGB-D data
acquired with a hand-held camera.  Although camera poses can usually
be tracked over short distances \cite{Newcombe11}, local tracking
often fails in texture-less regions and/or drifts over long ranges
\cite{Dai16,Niessner13} (left side of Figure \ref{fig:teaser}).  These errors 
can be fixed with asynchronous or global optimizations based on detected loop 
closures \cite{Choi15,Henry10,Whelan15}.  However, finding loop closures is 
difficult without prior constraints in large real-world scans with multiple 
rooms and/or repeated structures.  In our experience, even state-of-the-art 
global registration methods produce warped surfaces and improbable structures 
in these cases \cite{Choi15}.

To address this issue, global refinement methods have been proposed
based on fitting structural models \cite{Li11, Ma16, Monszpart15} and/or 
aligning closest point correspondences \cite{Brown07, Yu15}.  However, these methods only succeed when the alignments provided as input are nearly correct. Otherwise, they may detect and amplify erroneous constraints found
in the misaligned inputs.

\begin{figure}[t!]
\centering
\includegraphics[width=\columnwidth]{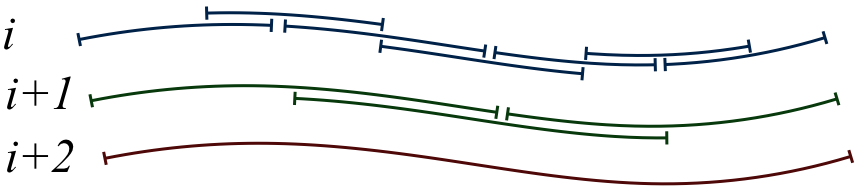} 
\caption{Schematic view of window sizes and their growth as iterations increase.  In every iteration, the camera trajectory is divided into multiple overlapping windows of equal length.  The window size is doubled in each iteration, until the entire trajectory is considered within a single window.}
\label{fig:windows}
\end{figure}

We introduce a new ``fine-to-coarse'' global registration algorithm that 
refines an initial set of camera poses by iteratively detecting and enforcing 
geometric constraints within gradually growing subsets of the scanned 
trajectory.  During each iteration, closest points and geometric constraints 
(plane parallelism, perpendicularity, etc.) are detected and enforced only 
within overlapping ``windows'' of sequential RGB-D frames.  The windows start
small, such that relative initial alignments are likely to be correct
(Figure~\ref{fig:windows}).  As the algorithm proceeds, the windows 
gradually increase in size, enabling detection of longer-range correspondences 
and large-scale geometric structures leveraging the improved trajectory
provided by previous iterations. This process is continued until final
window includes the entire scan and a global refinement can be done robustly.

\begin{figure*}[t!]
\centering
\includegraphics[width=2\columnwidth]{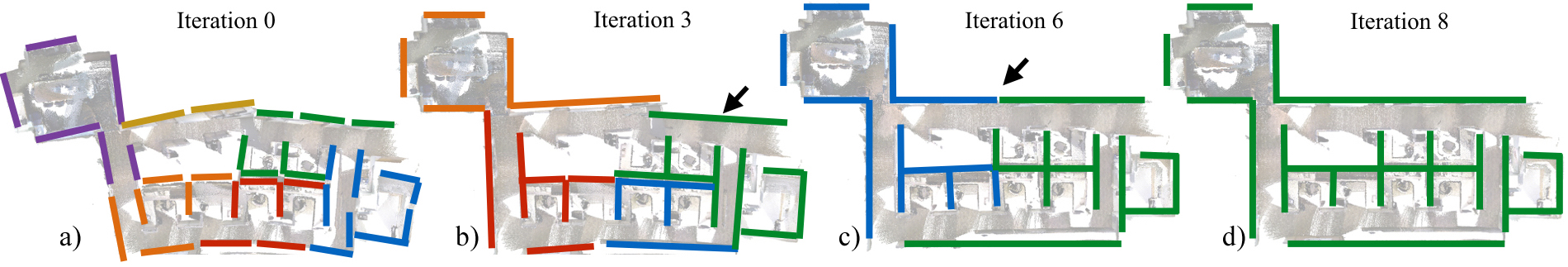}
\caption{Schematic view of fine-to-coarse registration. Starting with initial alignment $T_0$ shown on the left, our algorithm detects and enforces structures in local regions (color-coded) in the first few iterations. As the algorithm progresses, the trajectory is refined, allowing for detection of larger geometrical structures. By iteration 6, we have properly aligned the wall marked by the arrow, without using explicit loop closures. }
\label{fig:iterations}
\end{figure*}

The advantage of this fine-to-coarse approach is that closest point
correspondences and planar structures are detected in each iteration only at 
the scales at which previous iterations have already aligned the scans.  
Enforcing these constraints in one iteration improves the registration for the 
next.  For example in Figure \ref{fig:iterations}, note how geometric 
constraints between walls become easier to detect in each iteration (left to 
right), and enforcement of those constraints gradually rectifies the 
reconstruction.  In the final iteration, the alignment is almost perfect, 
making it trivial to detect very large-scale structures and long-range 
constraints (e.g., parallel walls in different rooms), which are crucial for 
correct global registration.  

To evaluate this algorithm and enable comparisons between future algorithms, we 
have created a new registration benchmark based on the SUN3D dataset \cite{
Xiao13b}.  It contains 10,401 manually-clicked point correspondences in RGB-D 
scans containing 149011 frames in 25 scenes, many of which span multiple rooms. 
During experiments with this new benchmark, we find that our fine-to-coarse 
algorithm produces more accurate global registrations and handles more 
difficult inputs than previous approaches.

Overall, the research contributions of this paper are three-fold.  First,
we propose a new fine-to-coarse, iterative refinement strategy for global
registration of large-scale RGB-D scans.  Second, we introduce a 
a new benchmark dataset for evaluating global registration algorithms 
quantitatively on real RGB-D scans.  Finally, we provide 
results of ablation studies revealing trade-offs for different components of a 
global registration algorithm.  The code and data for all three will be made 
publicly available.

\section{Related Work}
\label{sec:related}

There has been a long history of research on registration of RGB-D
images in both computer graphics and computer vision, as well as in augmented reality, robotics, and other fields \cite{Stotko16}. The following paragraphs describe the work most closely related to ours.

\hd{Real-time reconstruction.} Most prior work has focused on real-time registration motivated by SLAM applications in robotics and augmented reality \cite{Stotko16}. Early systems use ICP to estimate pairwise alignments of adjacent video frames \cite{Besl92} and feature matching techniques to detect and align loop closures \cite{Angeli08}.  More recent methods have aligned frames to a scene model, represented as a point cloud \cite{Henry10,Keller13,Rusinkiewicz02,Whelan15} or an implicit function \cite{Chen13,Dai16,Kahler15,Newcombe11,Wang16,Whelan12,Whelan14}. With these methods, small local alignment errors can accumulate to form gross inconsistencies at large scales \cite{Klingensmith_2015,Niessner13}.  

\hd{Off-line global registration.} To rectify misalignments in on-line
camera pose estimates, it is common to use off-line or asynchronously executed
global registration procedures.  A common formulation is to compute a pose graph with edges representing pairwise transformations between frames and then optimize an objective function penalizing deviations from these pairwise alignments \cite{Grisetti10,Henry10,Zhou13,Zhou14}.  A major challenge in these approaches is to identify which pairs should be connected by edges (loop closures).  Previous methods have searched for similar images with Bag-of-Words models \cite{Angeli08}, randomized fern encodings \cite{Whelan15}, convolutional neural networks \cite{Chen14a}, and other methods.  Choi et al.\cite{Choi15} recently proposed a method that uses indicator variables to identify true loop closures during the global optimization using a least-squares formulation.  In our experiments, their algorithm is successful on scans of small environments, but not for ones with multiple rooms, large-scale structures, and/or many repeated elements.

\hd{Hierarchical graph optimization.} Some methods fuse subgraphs of a pose graph hierarchically to improve optimization robustness and efficiency \cite{Choi15,Estrada05,Frese05,Ratter15,Tang15}.  Some of the ideas motivating these methods are related to ours.   However, they detect all potential loop closures before the optimization starts.  In contrast, we detect new constraints (planar relationships and feature correspondences) in the inner loop of an iterative refinement, which enables gradual discovery of large-scale structures and long-range constraints as the registration gets better.

\hd{Iterative refinement.} Other methods have used Iterative Closest Point (ICP) \cite{Besl92} to compute global registrations \cite{Pulli99, Brown07, Yu15}.   The advantage of this approach is that dense correspondences (including loop closures) are found only with local searches for closest points based on a prior alignments, rather than with global searches that consider all pairs of frames.  However, ICP generally requires a good initial alignment and thus is rarely used for global RGB-D registration except as fine-scale refinement in the last step \cite{Choi15}.  Our work addresses that limitation.

\section{Approach}
\label{sec:Overview}

In this paper, we describe a global registration algorithm that
leverages detection and enforcement of nearly-satisfied constraints in
the inner loop of an iterative registration refinement.  The algorithm
starts with an initial registration and then follows the general E-M
strategy of alternating between a discrete E step (detecting
constraints) and a continuous M step (solving for the camera poses
that best satisfy the constraints).

Though the method is general, we consider two types of constraints in
this work: feature correspondences and planar structure relationships.
During each iteration of the algorithm, constraints are created based
on correspondences between closest compatible features (like in ICP)
and based on geometric relationships (coplanarity, parallelism, orthogonality, etc.) between detected planar structures.  The
constraints are integrated into a global optimization that refines
camera poses before proceeding to the next iteration.

The key new idea is that the detection of constraints occurs only within 
windows of sequential RGB-D frames that grow gradually as the iterative 
algorithm proceeds.  In the early iterations, the 
windows span just a few sequential RGB-D frames where relative camera poses of 
the initial alignment should be nearly correct.  At this early stage, for 
example, it should be possible to detect coplanarity and orthogonality 
constraints between nearby surfaces in adjacent frames (Figure~\ref
{fig:iterations}a).  As the iterations proceed, the windows get larger, 
enabling detection and enforcement of larger-scale and longer-range feature
correspondences and planar constraints (Figure~\ref{fig:iterations}c).
At each iteration, we expect the relative camera poses within each
window to be approximately correct, since they have been optimized
according to constraints detected in smaller windows during
previous iterations.  Thus, we expect that it will be possible to
detect the relevant structural constraints within each window robustly
based on the current camera pose estimates.  Ultimately, in the final
iteration, the algorithm uses a single window encompassing the entire
scan.  At that stage, it detects and enforces a single structural
model within a global optimization of all camera poses (Figure~
\ref{fig:iterations}d).

We call this method ``fine-to-coarse'' because it is the 
opposite of ``coarse-to-fine:'' alignment at the fine-scale
facilitates alignment at the coarse-scale, rather than vice-versa.

The main advantages of this approach are two-fold.  First, it avoids a
global search for pairwise loop closures -- they are instead found
incrementally as they become nearly aligned.  Second, it enables
detection and enforcement of large-scale geometric constraints (like
planar structure relationships) even though they might not be evident
in the initial alignment (e.g., the parallel relationship between the
leftmost and rightmost walls in the example of Figure
\ref{fig:iterations} would be difficult to infer in iteration 0, but
is simple to detect in Iteration 6).  As a result, our method achieves significantly better registration results for large-scale scans compared to previous methods (Section~\ref{sec:evaluation}).

\section{Algorithm}
\label{sec:algorithm}

The input to our system is a set of $n$ RGB-D images $I$ acquired with a consumer level RGB-D camera.  The output is set of camera poses $T$, where $T[k]$ represents the position and orientation of the camera for $I[k]$.

Processing proceeds as shown in Algorithm~\ref{fig:pseudocode}.
During a preprocessing phase, we first extract features $F$ and planar regions 
$P$ from all images in $I$, estimate pairwise local alignment transformations $L
[k]$ for successive images $I[k-1]$ and $I[k]$, and concatenate local 
transformations to form an initial guess for global transformations $T_0$.   
Then, we iteratively refine the global transformations $T_i$ by detecting a set 
of planar proxies $P_i$, creating structural constraints $S_i$ between those 
proxies, detecting feature correspondence constraints $C_i$, and 
optimizing the global transformations for the next iteration $T_{i+1}$ by 
minimizing an error function encoding those constraints.  At each iteration $i$,
constraints are detected and enforced only within windows of consecutive $l_i$ 
images, with each window $W_{i}[j]$ overlapping its neighbors by $l_{i}/2$.  
The size of the windows grows by a factor of two after each iteration, 
providing fine-to-coarse alignment, until they cover the entire set of $n$ 
input images.
The following subsections describe the core ideas for each of these steps.  The 
full implementation details appear in the supplemental material.

\begin{algorithm}[t!]
 \textbf{Input}: Images {$I$, window length $l_0$, $n\_iter$}\;
 \textbf{Output}: Camera transformations {$T$}\;
 $F$ = ExtractFeatures($I$)\;
 $P_0$ = CreateProxies($I$)\;
 $L[k]$ = AlignImages($I[k-1]$, $I[k]$)\;
 $T_0$ = Concatenate($L[i]$)\;
 \For{$i \leftarrow 0$ \KwTo $n\_iter$}{
  \lForEach{$W_i[j]$}{$P_i[j]$=CreateProxies($P$,$W_i[j]$)}
  \ForEach{$W_i[j]$}{ 
    $S_i[j]$=StructuralConstraints($P_i,T_i,j$)\;
    $C_i[j]$=CorrespConstraints($F,W_i[j],T_i$);
    }
  $T_{i+1}$=Solve $argmin_{T} E(C_i,S_i,T_i)$\; 
  $l_{i+1}=2 * l_i$\; 
}
\caption{Fine-to-coarse refinement}
\label{fig:pseudocode}
\end{algorithm}

\subsection{Preprocessing}
\label{sec:alg-init}

\hd{Extracting Features.} 
The first step of preprocessing is to extract a dense set of features $F[k]$ from each input depth image $I[k]$.  Our goal in this step is mainly to construct a set of well-spaced and repeatable features that can be matched robustly later when searching for closest correspondences.  We experimented with a number of feature types, including SIFT and Harris corners in both color and depth images.   However, we ultimately found planar patches \cite{Bartoli03,Weingarten06,Nguyen07,Pathak10,Salas-Moreno14,Dou12,Trevor12,Elghor15,Taguchi13,Ma16,Dzitsiuk16} and linear edges along creases and contours in depth images \cite{Zhou15a} to be most robust, and so we use only them.  

\hd{Creating Planar Proxies.} 
The next step is to extract planar proxies from each depth image.  The goal here is to extract planar regions that can form the basis of structural constraints introduced later during the fine-to-coarse iterations.  To do this, we use a method based on agglomerative clustering, which we found to perform well on the noisy depth data tested in this study.  

\hd{Aligning Adjacent Images.}
The following step is to estimate a local alignment transformation $L[k]$ for each pair of successive input images, $I[k-1] \rightarrow I[k]$.  Our goal in this step is to provide a local alignment transformation that can be used later in the optimization to preserve the local shape of the estimated camera trajectory.   To do this, we use a pairwise image alignment approach based on Xiao et al. \cite{Xiao13b}: we detect SIFT features in color images, prune out ones without valid depth values, assign 3D positions to the rest, and then use RANSAC in 3D to search for the rigid transformation $L[k]$ aligning as many of these features as possible.  

\hd{Initializing Transformations.}
The final pre-computation step is to estimate an initial transformation $T_0$[k]
for each image $I[k]$.  Since these transformations are input to our 
fine-to-coarse algorithm, they do not need to be globally consistent.  Rather 
it is important that they are locally accurate.  So, we form $T_0[k]$ by simply 
concatenating local transformations in $L$: ($T_0[0]=I$; $T_0[k]=L[k-1]T_0[k-1]$; $k \in [1,n]$).  

\subsection{Fine-to-Coarse Refinement}
\label{sec:alg-opt}

After preprocessing the images, the algorithm iteratively detects constraints within windows of increasing sizes and solves for camera transformations based on those constraints.  The input to each iteration $i$ is a window size $l_i$ and a set of camera transformations $T_i$ from the previous iteration.  The output is a set of new camera transformations $T_{i+1}$, computed as follows:

\hd{Creating Co-planarity Constraints.} 
We first build a set of co-planarity constraints $H$ relating proxies in $P$ for each overlapping window of successive images.  We do that by computing a global position for every planar proxy in $P$ using the current transformations $T_i$. Then, for every window $W_{i}[j]$, we cluster co-planar proxies associated with features of images within the window using an agglomerative hierarchical clustering algorithm.   Each coplanar set constitutes a new planar proxy $P_l$ that is inserted into $P$, and co-planar, parent-child constraint $\{P_l,P_j ; j \in [1, |children(P_l)|]\}$ is inserted into $H$.

\begin{figure}[t]
\centering
\includegraphics[width=0.7\columnwidth]{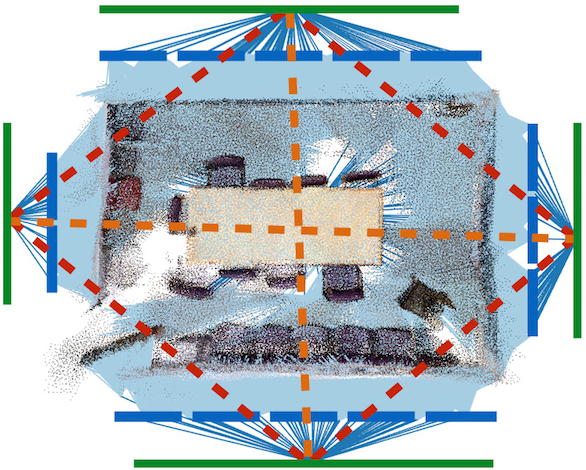}

\caption{Exploded view of our structural model for one of the SUN3D scenes. Geometrical properties like parallelism (dashed orange) and orthogonality (dashed red) are created between parent proxies (green). Parent proxies are connected to the scan features (point-cloud) through children proxies via co-planarity constraints (blue and light blue, respectively). }
\label{fig:structural_model}
\end{figure}

The constraints implied by $H$ are depicted for a single-room example in Figure \ref{fig:structural_model}.  Note that shown is a late in the fine-to-coarse refinement, and thus the planar structures in the green span entire walls.  In contrast to previous methods based on alignment to planes \cite{Ma16,Salas-Moreno14,zhang2015online}, it is possible to detect these large planar structures in our approach because previous iterations align overlapping subsets of the wall.

\hd{Creating Planar Relationship Constraints.} 
We next build a set of structural constraints $G$ representing geometric relationships between parent planar proxies in the same or adjacent windows.  Our goal is to detect salient relationships between planar structures (parallel, antiparallel, or orthogonal) that can help guide the optimization towards the correct registration.

We create a typed and weighted planar relationship for every pair of parent planar proxies ($P_a = \{\vec{n}_a, p_a\}$ and $P_b = \{\vec{n}_b, p_b\}$) within the same window.  Specifically, the type of the structural relationship $s_{ab}$ and its weight $w_{ab}$ are based on the angle $\theta=acos(\vec{n}_a, \vec{n}_b)$ between the normals.  
For parallel relationships the weight is defined as $w_{ab}=\exp(-\theta^2 / 2\sigma^2_\theta)$, for orthogonal $w_{ab}=\exp(-(\theta-\frac{\pi}{2})^2 / 2\sigma^2_\theta)$, and for antiparallel $w_{ab}= exp(-(\theta-\pi)^2 / 2\sigma^2_\theta)$.  These weights are chosen so as to guide the registration when constraints are nearly met, but have little influence when they are not.

\hd{Creating Feature Correspondences.} 
We next build a set of correspondence constraints $C$ between features detected in images within the same window.  Following the general strategy of ICP, we construct correspondences between the closest compatible features for every pair of scans in every $W_i[j]$, where compatibility is determined by a maximum distance threshold, a maximum normal angle deviation, and a feature type check (planar features only match to planar features, etc.).  No shape descriptors are used.

Since we expect images within the same window to become aligned as they are optimized, we set the maximum distance and maximum normal angle thresholds for rejecting outliers dynamically for every pair of images based on how many iterations they have been within the same window.  The first time the two images are paired, the thresholds are quite large: 0.5m and 45 degrees.  As the iterations proceed, they decrease linearly down to 0.15m and 20 degree.

Finally, for performance reasons, we subsample the set of correspondences created for all windows $W_i[j]$, such that the total number of them is equal to $|C| = 100n$.

\subsection{Optimization}
\label{sec:optimization}

The final processing step in each iteration is to optimize the camera transformations $T_i$ and planar proxies $P_i$ to minimize an error function encoding the detected constraints.

Our error function is a weighted sum of terms penalizing deformations of structural relationships ($E_S$), warps of detected planes ($E_P$), distances between corresponding features ($E_C$), misalignments of local transformations ($E_L$), and large changes in transformations ($E_I$):

\begin{equation*}
\begin{split} 
  E(T,S,C) = 
  & w_S E_S(S) + w_C E_C(T,C) \\
  & w_L E_L(T) + W_I E_I(T)
\end{split}
\label{eq:error}
\end{equation*}

\hd{Planar Structure Error.} $E_S$ is designed to encourage consistency between
the registration and the the structural model:
$$ E_S(T,S) = w_P E_P(T,P) + w_H E_H(H) + w_g E_G(G) $$
where the three sub-terms account for coplanarity of hierarchical proxy 
relationships ($E_H$), detected geometric relationships between proxies ($E_G$),
and fits of planar proxies to depth images ($E_P$).
The error due to coplanarity constraints is defined as:
\begin{equation*}
E_{H}(P) =\sum_{j=1}^{|H|} E_{cp}(P_k,P_j)
\end{equation*}
where $E_{cp}(A,B)$ measures the deviation of two planar structures from coplanarity.
The error in geometric relationships is:
\begin{equation*}
E_G(G) = \sum_{j=1}^{|G|}
\begin{cases}
w_{jk}(\vec{n}_j - \vec{n}_k)^2 & parallel \\
w_{jk}(\vec{n}_j + \vec{n}_k)^2 & antiparallel \\
w_{jk}(\vec{n}_j \cdot \vec{n}_k)^2 & orthogonal 
\end{cases}
\end{equation*}
where the $\vec{n}_j$ and $\vec{n}_k$ are normals of proxies $P_j$, $P_k$ respectively.
The error aligning planar proxies with depth images is:
\[
E_{P}(T,P) = \sum_{j=1}^{|P|} \sum_{k=1}^{m_j} E_{cp}(P_j,T[ik](F[k]))
\]
where $ik$ is the index of the image containing feature $F[k]$ and $m_j$ is a number proportional to inliers of proxy $P_j$, selected such that $\sum{m_j} = 100n$.

\hd{Feature Correspondence Error.}
$E_c$ is designed to encourage alignment of detected feature correspondences:
\begin{equation*}
E_{C}(T,C) =\sum_{j=1}^{|C|} 
\begin{cases}
((p_b - p_a) \times \vec{n}_a)^2 & edges \\
((p_b - p_a) \cdot \vec{n}_a)^2 & planes \\
\end{cases}
\end{equation*}
where $p_a$, $p_b$ and $\vec{n}_a$, $\vec{n}_b$ are the transformed positions  
and normals of features $F[a]$ and $F[b]$.

\hd{Local Alignment Error.} $E_L$ is designed to encourage pairwise transformations
between adjacent frames to match the ones
computed during preprocessing:
\begin{equation*}
\begin{split}
&E_L(T) = \sum_{j=0}^{n-1} \sum_{k=0}^{k_{max}} \\
&E_t( T_0[j+2^k]^{-1}(T_0[j]), T[j+2^k]^{-1}(T[j]) )
\end{split}
\end{equation*}
where $k_{max}=4$ and $E_t$ measures the misalignment of two transformations.

\hd{Inertia Error.} $E_I$ is designed to encourage the transformations to remain approximately the same across iterations.   
$$
E_{I}(T,P) =\sum_{j=1}^{|I|} (\Delta T[j])^2 + \sum_{j=1}^{|P|} (\Delta P[j])^2
$$
where $\Delta A$ represents the sum of squared differences between
Euler angle rotations and translations for $A$ from one iteration to the next.
This term has very low weight -- it is added mainly to provide stability for the optimization and prevent the system of equations from being under-constrained.

\section{Experimental Results}
\label{sec:evaluation}

We performed a series of experiments designed to test the
performance of the proposed method with comparisons to previous
methods and ablation studies. 

\hd{New Benchmark Dataset.} RGB-D scans of indoor scenes with ground
truth alignments are scarce.  Most contain only part of a room
\cite{Anand12,Dai16,Handa14,Lai14,Mattausch14,Silberman12,Sturm12},
have less than ten test examples \cite{Dai16,Mattausch14,Xiao13b}, or
are based on synthetic data \cite{Handa14,Handa16}. As a result, the
research community has compared registration results mainly on small,
clean datasets that are not representative of the large real-world
scans required for most applications.

\begin{figure}[t]
\centering
\includegraphics[width=\columnwidth]{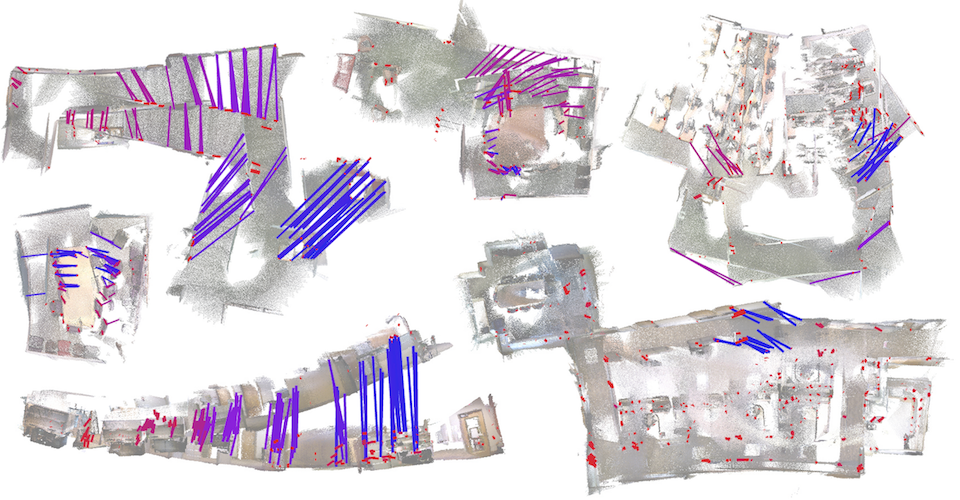}
\caption{Ground truth correspondences for 6 out of 25 scenes in our benchmark.  The visualization shows lines between manually-clicked corresponding points after alignment with $T_0$, the initial registration for our method. Color indicates the frame distance - blue correspondences denote loop closure pairs, while red denotes local pairs.}
\label{fig:groundtruth}
\end{figure}

\begin{figure*}[t!]
\centering
\includegraphics[width=\textwidth]{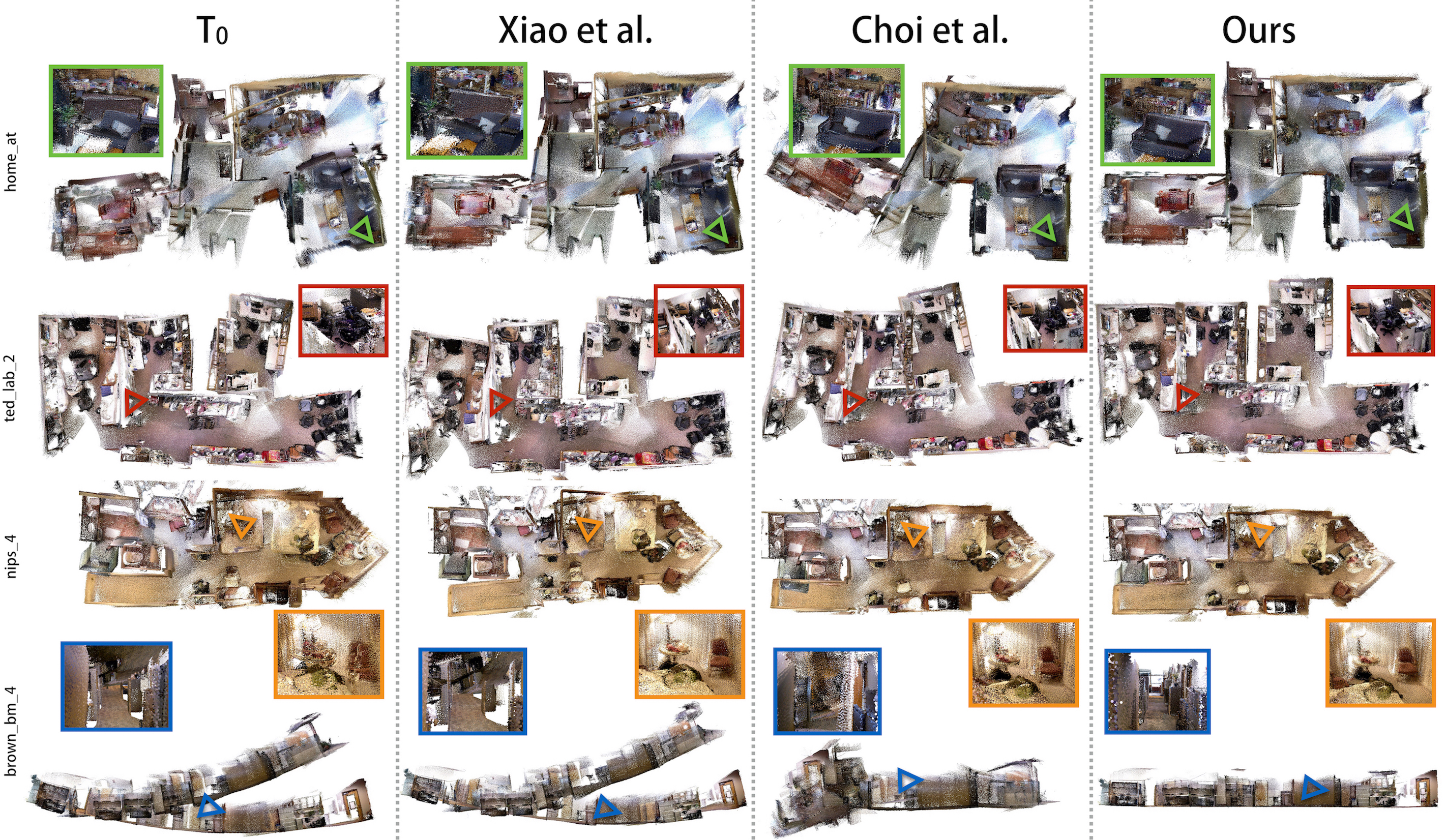}
\caption{Qualitative comparison of global registration results for example SUN3D scenes.  The rightmost column shows our results. The leftmost column shows the solution used to initialize our algorithm ($T_0$).  The middle two columns show results produced with prior work \cite{Choi15,Xiao13b}.  In insets, we show close-ups of particular regions.  In the first two rows, our method is able to recover correct arrangement of captured multi-room environments, while previous work produces improbable structures, like intersecting rooms.  The third row shows a sequence with non-Manhattan walls, which we are able to register correctly.  Our method is also able to correctly align a challenging corridor sequence in the fourth row, where for Xiao et al., the visual place recognition has failed.  Due to a lot of geometrical self similarities, Choi et al. is unable to recover proper geometry.}
\label{fig:images}
\end{figure*}

To address this issue, we introduce a new registration benchmark based
on the SUN3D dataset \cite{Xiao13b}.  SUN3D contains a large set RGB-D videos
captured with a ASUS Xtion PRO LIVE sensor attached to a hand-held
laptop in a variety of spaces (apartments, hotel rooms, classrooms,
etc.).  Each scan contains $10^3-10^4$ images, often covering multiple
rooms.  Previously, only eight of the scenes were released with full annotations and pose correction. Because of the lack of ground truth poses, these have not been used for quantitative evaluation of registration algorithms.

One of our contributions is to provide ground-truth point correspondences
for 25 of the largest scenes in SUN3D.   In all, we have manually clicked on
10,401 point correspondences with pixel-level accuracy.
These ground-truth correspondences are largely in pairs of overlapping frames 
forming loop closures, but they also appear in pairs of nearby frames spread 
evenly through the scan, as shown in Figure \ref{fig:groundtruth}. The average 
number of correspondences per scan is 416, with a minimum of 239 and a maximum 
of 714.

We use these ground truth correspondences to evaluate and compare RGB-D registration algorithms by computing their root mean squared error (RMSE).
To quantify a lower bound on the RMSE in this test, we aligned the ground truth correspondences for all scene with no other constraints and
report the errors in the left column of Table \ref{tab:results}.   Note that these lower-bounds are non-zero,
even though clicked correspondences are pixel-accurate,
due to the extreme noise in uncalibrated SUN3D depth maps.

\hd{Comparisons to Previous Work.}
We evaluate our method in comparison to two prior methods for offline 
registration: Xiao et al.'s Sun3DSfm\cite{Xiao13a} and Choi et al.'s
Robust Reconstruction of Indoor Scenes \cite{Choi15} (Figure~\ref{fig:images}). The first method by Xiao et al. uses the same method as we do 
for tracking camera poses, but also predicts loop closures via visual place 
recognition with a BoW approach and performs a global bundle adjustment to 
optimize for camera poses.  The second method by Choi et al. fuses consecutive 
groups of 50 frames into fragments, aligns all pairs of fragments with a 
variant of RANSAC, selects pairs as potential loop closures, and then solves a 
least squares system of nonlinear equations that simultaneously solves for 
camera poses and loop closure weights.  We believe this second method is the 
state-of-the-art for off-line global registration amongst ones with code 
available, even though it only uses the depth information.  Comparisons are 
provided in the supplemental materials for several real-time reconstruction 
methods, which demonstrate worse performance than these off-line global methods.

\begin{table}[t]
\centering
\resizebox{\columnwidth}{!}{%
\begin{tabular}{r|c|c|c|c|c|}
\cline{2-6}
                                         & Ground Truth & Ours & $T_0$ & Xiao et al. & Choi et al. \\ \hline
\multicolumn{1}{|l|}{Average}            & 0.031 & 0.077 & 0.531 & 0.431 & 0.993 \\
\hline
\multicolumn{1}{|l|}{Standard Deviation} & 0.006 & 0.037 & 0.471 & 0.491 & 1.481 \\
\hline
\multicolumn{1}{|l|}{Median}             & 0.031 & 0.065 & 0.421 & 0.255 & 0.224 \\
\hline
\multicolumn{1}{|l|}{Minimum}            & 0.019 & 0.042 & 0.114 & 0.078 & 0.043 \\
\hline
\multicolumn{1}{|l|}{Maxium}            & 0.045 & 0.221 & 2.124 & 2.007 & 5.819 \\
\hline
\end{tabular}
}
\vspace{2mm}
\caption{Comparison of RMSE statistics in meters with different registration methods for the 25 scenes in our SUN3D benchmark.}
\label{tab:results}
\end{table}

\begin{figure}[t]
\centering
\vspace*{-0.1in}
\includegraphics[width=\columnwidth]{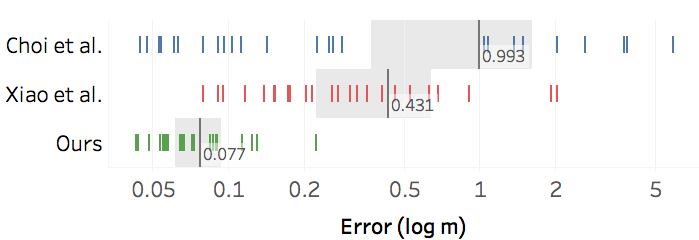}
\caption{Quantitative comparison.  Every vertical bar in each row represents the RMSE achieved for one of the 25 SUN3D scenes with the algorithm listed on the left.  The vertical gray bar shows the average RMSE for each method, and the shaded gray regions represents one standard deviation.}
\label{fig:results}
\end{figure}

Table \ref{tab:results} and Figure \ref{fig:results} show quantitative results 
for the comparison evaluated on our new SUN3D benchmark.  Table \ref
{tab:results} compares overall statistics of RMSEs for each algorithm, while 
Figure \ref{fig:results} shows the distributions of RMSEs.   It can be seen in 
both of these results that our reconstruction algorithm aligns the ground truth 
correspondences better than either of the other two methods: our median error 
is 0.065m in comparison to 0.255m for Xiao et al. and 0.224m for Choi et al.  
In case-by-case comparisons, our method has the lowest error in 21 of 25 scenes.  

\begin{figure}[t]
\centering
\includegraphics[width=\columnwidth]{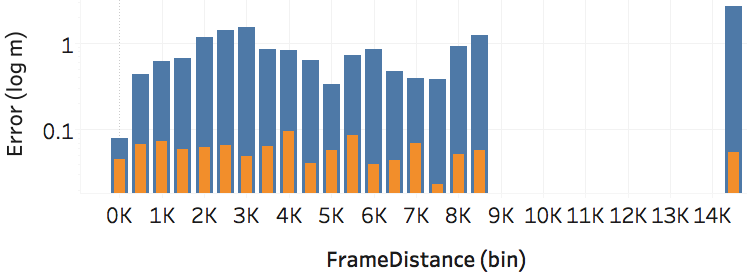}
\caption{Investigating fine-to-coarse iteration.  Each bin gathers 
correspondences that are specific numbers of frames away from each other in the 
RGB-D video. Blue bars show the correspondence errors using initial pairwise 
transformations ($T_0$), while orange bars show errors after applying our method (on a log scale).
Note that errors decrease for both long-range loop closures and 
nearby frames.}
\label{fig:distance}
\end{figure}

\hd{Investigating Fine-to-Coarse Iteration.}
To investigate the behavior of our fine-to-coarse algorithm, we computed 
histograms of $L_2$ distances versus frame index differences between pairs of  
frames linked by ground-truth correspondences.  Figure \ref{fig:distance} shows 
a comparison of these histograms for the registrations at the start of our 
algorithm (blue) and at the end (orange).  It is interesting to note that our 
algorithm not only reduces the distances between ground-truth correspondences 
forming long-range loop closures (the right side of the plot), but also over 
short ranges.  This demonstrates that the extracted structural model helps to 
fix not only global alignments, but also local ones.

\begin{figure}[h]
\centering
\includegraphics[width=\columnwidth]{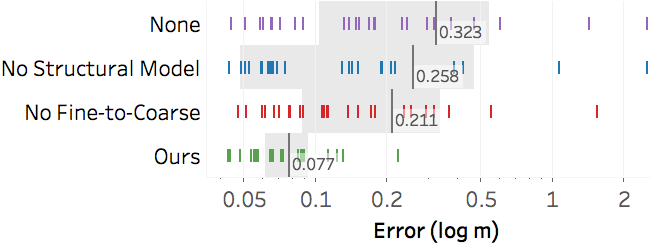}
\caption{Ablation studies.  Distributions of errors in the SUN3D benchmark for alternatives of our algorithm.  Disabling coarse-to-fine iteration or structural modeling diminishes performance. }
\label{fig:method_variants}
\end{figure}

\begin{figure}[t]
\centering
\includegraphics[width=\columnwidth]{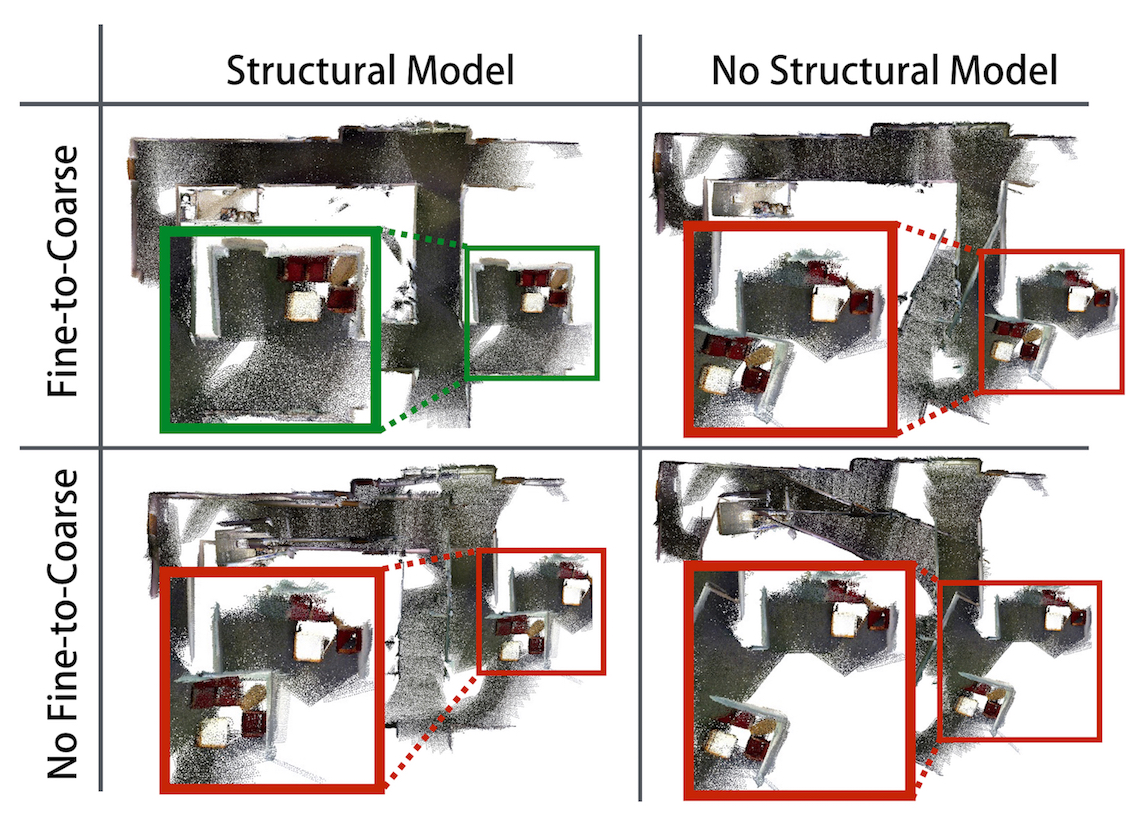}
\caption{Qualitative examples of our ablation studies. Only our full method, using both fine-to-coarse strategy and structural model is able to align the region with red chairs correctly (see zoom-in)}
\label{fig:method_variants_qual}
\end{figure}

\hd{Ablation Studies.}
To investigate the value of our proposed a) fine-to-coarse iteration strategy and b) structural model, we performed comparisons of our method with all combinations of these methods enabled or disabled.   The results in Figure \ref{fig:method_variants} and \ref{fig:method_variants_qual} show that both provide critical improvements to the results.   In particular, it is interesting to note that the fine-to-coarse iteration strategy improves ICP, even when no structural model is computed.   This result highlights the value of aligning local structures before searching for loop closures at larger scales.

\begin{figure}[t]
\centering
\includegraphics[width=\columnwidth]{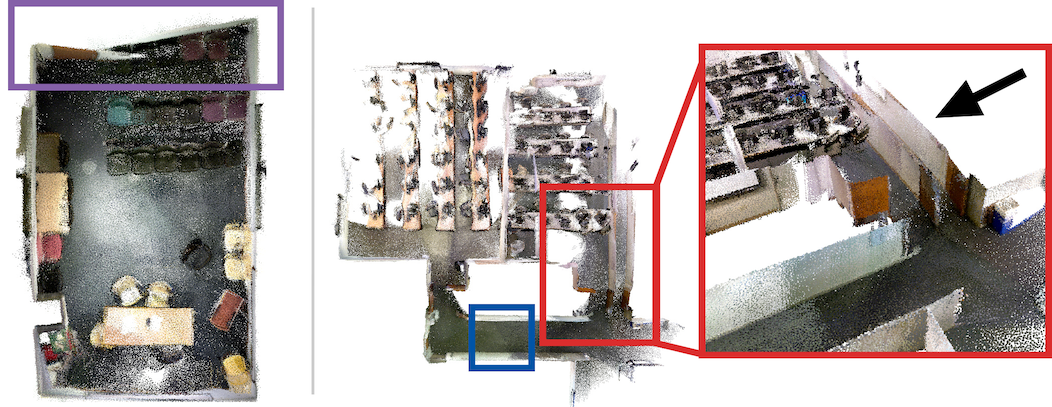}
\caption{Failure cases of our method. Left: the real world room is a trapezoid.
Our structural model introduces error, attempting to create a rectangular room. Right: Sliding along the corridor (in blue) causes failure in detecting loop closures.  Note the misaligned wall on the right marked by an arrow. }
\label{fig:failure_cases}
\end{figure}

\hd{Failure Cases.}
Our method does not always succeed.  For example, it can fail when multiple rooms are connected via featureless straight corridors and when rooms that are nearly (but not exactly) rectangular (Figure~\ref{fig:failure_cases}).  Failures of the second type are rare -- since the weight of enforcing parallelism and orthogonality constraints is low for pairs of planes at off-angles, we are able to reconstruct most scenes with non-Manhattan geometry correctly (as in the third row of Figure \ref{fig:results}).

\hd{Timing.}
All our tests took less than an hour to finish on a machine with 3.0GHz processor and 128Gb of RAM. The shortest sequence of 875 frames took 108 seconds to complete, while the longest took 2,909 seconds for 14,915 frames.

\section{Conclusions}
\label{sec:conclusions}
 
This paper describes a method for global registration of RGB-D scans captured with a hand-held camera in a typical indoor environment.  The key idea is a fine-to-coarse scheme that detects and enforces constraints (planar relationships and feature correspondences) within windows of gradually increasing scales in an iterative global optimization algorithm.  The benefits of the proposed approach are demonstrated in experiments with a new benchmark for RGB-D registration, which contains 10,401 manually specified correspondences across 25 SUN3D scenes.  This benchmark and all code will be publicly released to facilitate evaluation and comparison of future algorithms.


{\small
\bibliographystyle{ieee}
\bibliography{structured_registration}

\begin{thebibliography}{10}\itemsep=-1pt

\bibitem{Anand12}
A.~Anand, H.~S. Koppula, T.~Joachims, and A.~Saxena.
\newblock Contextually guided semantic labeling and search for 3d point clouds.
\newblock {\em IJRR}, 2012.

\bibitem{Angeli08}
A.~Angeli, D.~Filliat, S.~Doncieux, and J.-A. Meyer.
\newblock Fast and incremental method for loop-closure detection using bags of
  visual words.
\newblock {\em Robotics, IEEE Transactions on}, 24(5):1027--1037, 2008.

\bibitem{Bartoli03}
A.~Bartoli and P.~Sturm.
\newblock Constrained structure and motion from multiple uncalibrated views of
  a piecewise planar scene.
\newblock {\em International Journal of Computer Vision}, 52(1):45--64, 2003.

\bibitem{Besl92}
P.~J. Besl and N.~D. McKay.
\newblock A method for registration of {3-D} shapes.
\newblock {\em IEEE Trans. PAMI}, 14(2):239--256, 1992.

\bibitem{Brown07}
B.~Brown and S.~Rusinkiewicz.
\newblock Global non-rigid alignment of {3D} scans.
\newblock {\em {ACM} Transactions on Graphics (Proc. {SIGGRAPH})}, 26(3), Aug.
  2007.

\bibitem{Chen13}
J.~Chen, D.~Bautembach, and S.~Izadi.
\newblock Scalable real-time volumetric surface reconstruction.
\newblock {\em ACM Trans. Graph.}, 32(4):113:1--113:16, July 2013.

\bibitem{Chen14a}
Z.~Chen, O.~Lam, A.~Jacobson, and M.~Milford.
\newblock Convolutional neural network-based place recognition.
\newblock {\em arXiv preprint arXiv:1411.1509}, 2014.

\bibitem{Choi15}
S.~Choi, Q.-Y. Zhou, and V.~Koltun.
\newblock Robust reconstruction of indoor scenes.
\newblock In {\em IEEE Conference on Computer Vision and Pattern Recognition
  (CVPR)}, 2015.

\bibitem{Dai16}
A.~Dai, M.~Nie{\ss}ner, M.~Zollh{\"o}fer, S.~Izadi, and C.~Theobalt.
\newblock Bundlefusion: Real-time globally consistent 3d reconstruction using
  on-the-fly surface re-integration.
\newblock {\em arXiv preprint arXiv:1604.01093}, 2016.

\bibitem{Dou12}
M.~Dou, L.~Guan, J.-M. Frahm, and H.~Fuchs.
\newblock Exploring high-level plane primitives for indoor 3d reconstruction
  with a hand-held rgb-d camera.
\newblock In {\em Asian Conference on Computer Vision}, pages 94--108, 2012.

\bibitem{Dzitsiuk16}
M.~Dzitsiuk, J.~Sturm, R.~Maier, L.~Ma, and D.~Cremers.
\newblock De-noising, stabilizing and completing 3d reconstructions on-the-go
  using plane priors.
\newblock {\em CoRR}, abs/1609.08267, 2016.

\bibitem{Elghor15}
H.~E. Elghor, D.~Roussel, F.~Ababsa, and E.~H. Bouyakhf.
\newblock Planes detection for robust localization and mapping in rgb-d slam
  systems.
\newblock In {\em 3D Vision (3DV), 2015 International Conference on}, pages
  452--459, Oct 2015.

\bibitem{Estrada05}
C.~Estrada, J.~Neira, and J.~Tard´os.
\newblock Hierarchical slam: Real-time accurate mapping of large environments.
\newblock {\em Transactions on Robotics}, 21(4):588–596, 2005.

\bibitem{Yu15}
J.~X. Fisher~Yu and T.~Funkhouser.
\newblock Semantic alignment of lidar data at city scale.
\newblock In {\em 28th IEEE Conference on Computer Vision and Pattern
  Recognition}, 2015.

\bibitem{Frese05}
U.~Frese, P.~Larsson, and T.~Duckett.
\newblock A multilevel relaxation algorithm for simultaneous localisation and
  mapping.
\newblock {\em {IEEE} Transactions on Robotics}, 21(2):1–12, 2005.

\bibitem{Grisetti10}
G.~Grisetti, R.~Kümmerle, C.~Stachniss, and W.~Burgard.
\newblock A tutorial on graph-based slam.
\newblock {\em IEEE Intelligent Transportation Systems Magazine}, 2(4):31--43,
  2010.

\bibitem{Handa16}
A.~Handa, V.~Patraucean, V.~Badrinarayanan, S.~Stent, and R.~Cipolla.
\newblock Scenenet: Understanding real world indoor scenes with synthetic data.
\newblock In {\em IEEE CVPR}, 2016.

\bibitem{Handa14}
A.~Handa, T.~Whelan, J.~McDonald, and A.~Davison.
\newblock A benchmark for {RGB-D} visual odometry, {3D} reconstruction and
  {SLAM}.
\newblock In {\em IEEE Intl. Conf. on Robotics and Automation, ICRA}, Hong
  Kong, China, May 2014.

\bibitem{Henry10}
P.~Henry, M.~Krainin, E.~Herbst, X.~Ren, and D.~Fox.
\newblock Rgb-d mapping: Using depth cameras for dense 3d modeling of indoor
  environments.
\newblock In {\em International Symposium on Experimental Robotics (ISER)},
  2010.

\bibitem{Kahler15}
O.~Kahler, V.~A. Prisacariu, C.~Y. Ren, X.~Sun, P.~H.~S. Torr, and D.~W.
  Murray.
\newblock {Very High Frame Rate Volumetric Integration of Depth Images on
  Mobile Device}.
\newblock {\em {IEEE Transactions on Visualization and Computer Graphics
  (Proceedings International Symposium on Mixed and Augmented Reality 2015}},
  22(11), 2015.

\bibitem{Keller13}
M.~Keller, D.~Lefloch, M.~Lambers, S.~Izadi, T.~Weyrich, and A.~Kolb.
\newblock Real-time 3d reconstruction in dynamic scenes using point-based
  fusion.
\newblock In {\em 2013 International Conference on 3D Vision~-- 3DV}, pages
  1~-- 8, DOI:10.1109/3DV.2013.9, June 2013.

\bibitem{Klingensmith_2015}
M.~Klingensmith, I.~Dryanovski, S.~Srinivasa, and J.~Xiao.
\newblock Chisel: Real time large scale 3d reconstruction onboard a mobile
  device.
\newblock In {\em Robotics Science and Systems 2015}, July 2015.

\bibitem{Lai14}
K.~Lai, L.~Bo, and D.~Fox.
\newblock Unsupervised feature learning for 3d scene labeling.
\newblock In {\em IEEE International Conference on Robotics and Automation},
  page 3050–3057, 2014.

\bibitem{Li11}
Y.~Li, X.~Wu, Y.~Chrysanthou, A.~Sharf, D.~Cohen-Or, and N.~J. Mitra.
\newblock Globfit: Consistently fitting primitives by discovering global
  relations.
\newblock {\em ACM Transactions on Graphics}, 30(4):52:1--52:12, 2011.

\bibitem{Ma16}
L.~Ma, C.~Kerl, J.~Stueckler, and D.~Cremers.
\newblock Cpa-slam: Consistent plane-model alignment for direct rgb-d slam.
\newblock In {\em Int. Conf. on Robotics and Automation}, 2016.

\bibitem{Mattausch14}
O.~Mattausch, D.~Panozzo, C.~Mura, O.~Sorkine-Hornung, and R.~Pajarola.
\newblock Object detection and classification from large-scale cluttered indoor
  scans.
\newblock {\em Computer Graphics Forum}, 33(2):11–21, 2014.

\bibitem{Monszpart15}
A.~Monszpart, N.~Mellado, G.~J. Brostow, and N.~J. Mitra.
\newblock Rapter: Rebuilding man-made scenes with regular arrangements of
  planes.
\newblock {\em ACM Trans. Graph.}, 34(4):103:1--103:12, July 2015.

\bibitem{Newcombe11}
R.~A. Newcombe, A.~J. Davison, S.~Izadi, P.~Kohli, O.~Hilli\~ges, J.~Shotton,
  D.~Molyneaux, S.~Hodges, D.~Kim, and .~A. Fitzgibbon.
\newblock Kinectfusion: Real-time dense surface mapping and tracking.
\newblock In {\em Mixed and augmented reality (ISMAR), 2011 10th IEEE
  international symposium on}, pages 127--136. IEEE, 2011.

\bibitem{Nguyen07}
V.~Nguyen, A.~Harati, and R.~Siegwart.
\newblock A lightweight slam algorithm using orthogonal planes for indoor
  mobile robotics.
\newblock In {\em Intelligent Robots and Systems, 2007. IROS 2007. IEEE/RSJ
  International Conference on}, pages 658--663. IEEE, 2007.

\bibitem{Niessner13}
M.~Nie{\ss}ner, M.~Zollh\"ofer, S.~Izadi, and M.~Stamminger.
\newblock Real-time 3d reconstruction at scale using voxel hashing.
\newblock {\em ACM Transactions on Graphics (TOG)}, 2013.

\bibitem{Pathak10}
K.~Pathak, A.~Birk, N.~Vaskevicius, and J.~Poppinga.
\newblock Fast registration based on noisy planes with unknown correspondences
  for {3-D} mapping.
\newblock {\em {IEEE} Trans. Robotics}, 26(3):424–441, June.

\bibitem{Pulli99}
K.~Pulli.
\newblock Multiview registration for large data sets.
\newblock In {\em Proceedings of the 2Nd International Conference on 3-D
  Digital Imaging and Modeling}, 3DIM'99, pages 160--168, Washington, DC, USA,
  1999. IEEE Computer Society.

\bibitem{Ratter15}
A.~Ratter and C.~Sammut.
\newblock Local map based graph slam with hierarchical loop closure and
  optimisation.
\newblock 2015.

\bibitem{Rusinkiewicz02}
S.~Rusinkiewicz, O.~Hall-Holt, and M.~Levoy.
\newblock Real-time {3D} model acquisition.
\newblock {\em ACM Transactions on Graphics (Proc. SIGGRAPH)}, 21(3):438--446,
  July 2002.

\bibitem{Salas-Moreno14}
R.~Salas-Moreno, B.~Glocken, P.~Kelly, and A.~Davison.
\newblock Dense planar slam.
\newblock In {\em Mixed and Augmented Reality (ISMAR), 2014 IEEE International
  Symposium on}, pages 157--164, Sept 2014.

\bibitem{Silberman12}
N.~Silberman, D.~Hoiem, P.~Kohli, and R.~Fergus.
\newblock Indoor segmentation and support inference from rgbd images.
\newblock In {\em Proc. European Conf. on Comp. Vision}, 2012.

\bibitem{Stotko16}
P.~Stotko.
\newblock State of the art in real-time registration of rgb-d images.
\newblock In {\em CESCG}, 2016.

\bibitem{Sturm12}
J.~Sturm, N.~Engelhard, F.~Endres, W.~Burgard, and D.~Cremers.
\newblock A benchmark for the evaluation of rgb-d slam systems.
\newblock In {\em Proc. of the International Conference on Intelligent Robot
  Systems (IROS)}, Oct. 2012.

\bibitem{Taguchi13}
Y.~Taguchi, Y.-D. Jian, S.~Ramalingam, and C.~Feng.
\newblock Point-plane slam for hand-held 3d sensors.
\newblock In {\em Robotics and Automation (ICRA), 2013 IEEE International
  Conference on}, pages 5182--5189, May 2013.

\bibitem{Tang15}
Y.~Tang and J.~Feng.
\newblock Hierarchical multiview rigid registration.
\newblock {\em Comput. Graph. Forum}, 34(5):77--87, Aug. 2015.

\bibitem{Trevor12}
A.~Trevor, J.~{Rogers III}, and H.~Christensen.
\newblock Planar surface {SLAM} with {3D} and {2D} sensors.
\newblock In {\em ICRA}, 2012.

\bibitem{Wang16}
H.~Wang, J.~Wang, and L.~Wang.
\newblock Online reconstruction of indoor scenes from rgb-d streams.
\newblock In {\em IEEE CVPR}, 2016.

\bibitem{Weingarten06}
J.~Weingarten and R.~Siegwart.
\newblock {3D SLAM} using planar segments.
\newblock In {\em IROS}, page 3062–3067, 2006.

\bibitem{Whelan12}
T.~Whelan, M.~Kaess, M.~Fallon, H.~Johannsson, J.~Leonard, and J.~McDonald.
\newblock Kintinuous: Spatially extended {K}inect{F}usion.
\newblock In {\em RSS Workshop on RGB-D: Advanced Reasoning with Depth
  Cameras}, Sydney, Australia, Jul 2012.

\bibitem{Whelan14}
T.~Whelan, M.~Kaess, H.~Johannsson, M.~Fallon, J.~Leonard, and J.~McDonald.
\newblock Real-time large scale dense {RGB-D SLAM} with volumetric fusion.
\newblock {\em Intl. J. of Robotics Research, IJRR}, 2014.

\bibitem{Whelan15}
T.~Whelan, S.~Leutenegger, R.~F. Salas-Moreno, B.~Glocker, and A.~J. Davison.
\newblock {ElasticFusion}: Dense {SLAM} without a pose graph.
\newblock In {\em Robotics: Science and Systems (RSS)}, Rome, Italy, July 2015.

\bibitem{Xiao13b}
J.~Xiao, A.~Owens, and A.~Torralba.
\newblock Sun3d: A database of big spaces reconstructed using sfm and object
  labels.
\newblock {\em Computer Vision, IEEE International Conference on},
  0:1625--1632, 2013.

\bibitem{Xiao13a}
J.~Xiao, A.~Owens, and A.~Torralba.
\newblock {SUN3D} database: Semantic {RGB-D} bundle adjustment with human in
  the loop.
\newblock In {\em Proc. Int. Conf. on Comp. Vision}, 2013.

\bibitem{zhang2015online}
Y.~Zhang, W.~Xu, Y.~Tong, and K.~Zhou.
\newblock Online structure analysis for real-time indoor scene reconstruction.
\newblock {\em ACM Transactions on Graphics (TOG)}, 34(5):159, 2015.

\bibitem{Zhou13}
Q.-Y. Zhou and V.~Koltun.
\newblock Dense scene reconstruction with points of interest.
\newblock {\em {ACM} Transactions on Graphics}, 32(4), 2013.

\bibitem{Zhou14}
Q.-Y. Zhou and V.~Koltun.
\newblock Color map optimization for 3d reconstruction with consumer depth
  cameras.
\newblock {\em SIGGRAPH Conf. Proc.}, 2014.

\bibitem{Zhou15a}
Q.-Y. Zhou and V.~Koltun.
\newblock Depth camera tracking with contour cues.
\newblock In {\em The IEEE Conference on Computer Vision and Pattern
  Recognition (CVPR)}, June 2015.

\end{thebibliography}
}

\end{document}